\documentclass[runningheads]{llncs}
\usepackage{graphicx}

\usepackage{amssymb}
\usepackage{tabularx}
\usepackage{array}
\usepackage{booktabs}
\newcolumntype{P}[1]{>{\centering\arraybackslash}p{#1}}

\usepackage{graphicx}
\usepackage{comment}

\usepackage{makecell}

\usepackage[most]{tcolorbox}
\usepackage{colortbl}
\usepackage{ulem}

\begin{document}
\title{CREPE: Coordinate-Aware End-to-End Document Parser}
%
%
\author{
Yamato Okamoto\inst{1,2}$^\ast$$^\dagger$ \orcidID{0009-0009-2153-3782}\and 
\\
Youngmin Baek\inst{1,2}$^\ast$$^\ddagger$ \orcidID{0000-0001-7001-4641}\and 
\\ 
Geewook Kim\inst{1}\orcidID{0009-0001-6713-3858}\and
Ryota Nakao\inst{1,2}\orcidID{0009-0001-6692-6952}\and
\\ 
DongHyun Kim\inst{1}\orcidID{0000-0001-9033-5231}\and
Moon Bin Yim\inst{1}\orcidID{0000-0002-7272-2198}\and
\\ 
Seunghyun Park\inst{1}\orcidID{0000-0002-8509-9163}\and 
Bado Lee\inst{1}\orcidID{0000-0003-4962-8977}
}
\authorrunning{Y.Okamoto et al.}
%
\institute{
NAVER Cloud AI, Seongnam-si, Gyeonggi-do, Korea \\
\email{okamoto.yamato.w15@kyoto-u.jp} \\
\email{\{youngmin.baek, gw.kim, dong.hyun\}@navercorp.com} \\
\email{\{moonbin.yim, seung.park, bado.lee\}@navercorp.com}
\and
LINE WORKS, Shibuya-city, Toyko, Japan \\
\email{nakao.ryota@line-works.com}
}
\maketitle              
\begingroup
\renewcommand\thefootnote{}\footnote{$^\ast$ Equal contribution.}
\renewcommand\thefootnote{}\footnote{$^\dagger$ Current Affiliation: CyberAgent, Inc.}%
\renewcommand\thefootnote{}\footnote{$^\ddagger$ Corresponding author.}%

\addtocounter{footnote}{-2}%
\endgroup

\begin{abstract}
In this study, we formulate an OCR-free sequence generation model for visual document understanding (VDU). Our model not only parses text from document images but also extracts the spatial coordinates of the text based on the multi-head architecture.
Named as \textbf{C}oordinate-awa\textbf{r}e \textbf{E}nd-to-end Document \textbf{P}ars\textbf{e}r \textbf{(CREPE)}, our method uniquely integrates these capabilities by introducing a special token for OCR text, and token-triggered coordinate decoding. We also proposed a weakly-supervised framework for cost-efficient training, requiring only parsing annotations without high-cost coordinate annotations. 
Our experimental evaluations demonstrate CREPE's state-of-the-art performances on document parsing tasks. 
Beyond that, CREPE's adaptability is further highlighted by its successful usage in other document understanding tasks such as layout analysis, document visual question answering, and so one. CREPE's abilities including OCR and semantic parsing not only mitigate error propagation issues in existing OCR-dependent methods, it also significantly enhance the functionality of sequence generation models, ushering in a new era for document understanding studies.


\end{abstract}

\section{Introduction} 
\begin{figure}[t]
\centering
\includegraphics[width=0.9\textwidth]{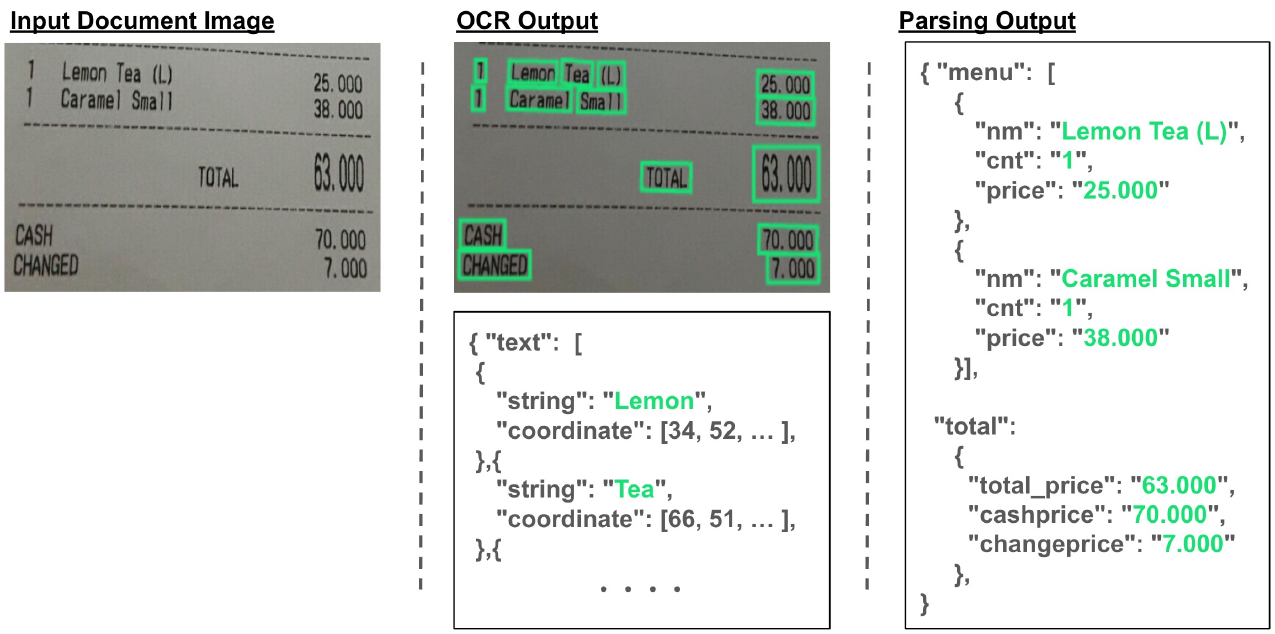}
\caption{
Examples of Parsing Task Outputs. OCR tasks aim to gather all text strings along with their corresponding coordinates in the document image, while parsing tasks aim to extract semantic information and present it in the required structured format.
}
\label{VDU_tasks}
\end{figure}

Visual Document Understanding (VDU), such as layout analysis, visual question answering, and document parsing, has evolved into a research field of growing interest in recent years~\cite{tanaka2024instructdoc}. In particular, document parsing, which aims to extract key information from document images and output it in a structured format as illustrated in Fig.~\ref{VDU_tasks}, is increasingly expected to facilitate the automation of processing a wide variety of documents.
Traditional parsing approaches use an off-the-shelf Optical Character Recognition (OCR) model to extract texts and their coordinates, and exploit the OCR outputs to parse the document images~\cite{xu2020layoutlm,xu2020layoutlmv2}.
To train these models, annotations for parsing tasks are necessary, including key-value relationships. Additionally, to avoid error propagation from OCR models, annotations for text strings and coordinates need to be provided, which can incur high-cost preparations.
Furthermore, utilizing OCR during inference may introduce additional limitations such as increased computation time, limited language support, and the upper limit of OCR performance.

To address these challenges, OCR-free end-to-end document understanding models like Donut~\cite{kim2022donut} and SeRum~\cite{Cao_2023_ICCV} have been proposed, which integrate text reading capability into parsing model. These methods not only simplify model development and management but also deliver notable performance in terms of computational cost-efficiency and parsing accuracy. Furthermore, they enable training models using only a dataset with parsing annotations, eliminating the need for costly annotations for all text strings and coordinates.
 
While OCR-free approaches have brought many benefits, previous methods also have limitations. One critical constraint is the challenge of providing text coordinates. Traditional OCR-based approaches can extract text coordinates as intermediate output thanks to the utilization of OCR models. In contrast, OCR-free approaches basically generate the parse texts sequentially, and are unable to provide text coordinates in their outputs.
Text coordinates play a crucial role in VDU applications. For example, some applications require masking sensitive information. Additionally, text position can be used for human verification processes in automated systems.

In this research, we aim to develop a model that simultaneously provides both parsing outputs and text coordinates, while retaining the advantages of OCR-free approaches, such as simple structure and independence of external OCR. We propose a new VDU model named CREPE (\textbf{C}oordinate-awa\textbf{r}e \textbf{E}nd-to-End Document \textbf{P}ars\textbf{e}r, and contributions are as follows:

\begin{enumerate}
    \item CREPE simultaneously generates parsing output along with the coordinates of text strings using the multi-head architecture.
    \item Without text coordinate annotations, CREPE can be trained on a dataset with annotations only for parsing, thanks to our newly proposed weakly supervised learning framework.
    \item CREPE can be applied to various applications that require image coordinates such as document layout analysis tasks, not only for parsing tasks.
\end{enumerate}

\section{Related Works} 

\subsection{VDU Tasks and Dataset}
\label{Related_VDUtasks}
Firstly, document parsing task, which is the main focus of our work, aims to extract key information such as named entities (e.g., company names, dates, addresses, total amounts), and output it in a structured format. This task involves understanding the semantic information of text and key-value relationships.
Popular datasets are FUNSD~\cite{jaume2019funsd}, which includes form category document images; SROIE~\cite{huang2019icdar2019} and CORD~\cite{park2019cord}, which contain receipt images; TrainTicket~\cite{guo2019eaten}, which contain real and synthetic train ticket images; and POIE~\cite{kuang2023visual}, which consists of images from nutrition facts labels of products.

Another task is document layout analysis, which involves recognizing and categorizing layout elements such as titles, figures, and tables in document images. Key datasets are PubLayNet~\cite{zhong2019publaynet} and DocBank~\cite{li2020docbank}, which consist of images of academic papers, and the ICDAR 2019 dataset~\cite{8978185} with complex document layouts.
Document visual question answering task aims to contextually answer questions about documents, which requires the model to interpret the document's layout and understand the textual content. A key dataset is DocVQA~\cite{mathew2021docvqa}.
Additionally, document image classification task aims to recognize the type of document. A notable dataset is RVL-CDIP~\cite{harley2015icdar}, featuring a variety of document images (e.g., invoices, letters, forms) and document category labels.

\begin{figure}[t]
\centering
\includegraphics[width=0.9\textwidth]{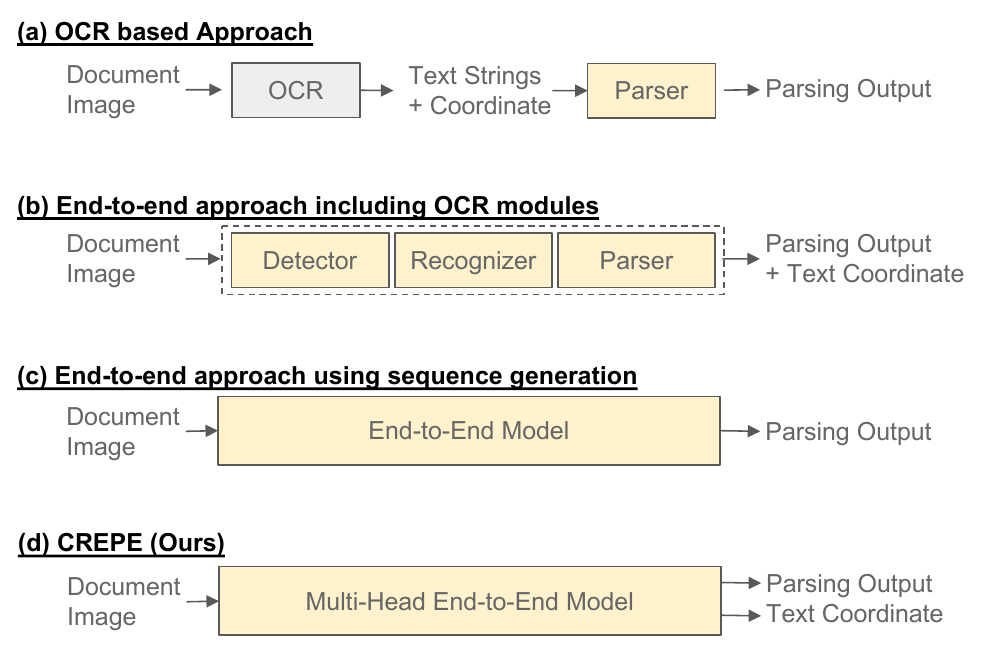}
\caption{
Differences among Traditional Approaches and Ours. (a) OCR-based methods leverage text strings and coordinates derived from an external OCR module. (b) End-to-end approach including OCR modules enable comprehensive optimization.  (c) End-to-end approaches using sequence generation decode the parsed texts without coordinates. (d) Our approach, termed CREPE, incorporates a multi-head architecture that preserves the advantages of an OCR-free method while uniquely enabling the provision of text coordinates.
}
\label{fig:related_work}
\end{figure}

\subsection{OCR-based Approach}
To understand real-world documents such as forms, receipts, and invoices, it is essential to consider both textual elements and visual elements, like format or layout. 
Therefore, several approaches use an off-the-shelf OCR to obtain both text strings and coordinates, and utilize them as textual features and visual features, respectively, as shown in Fig.~\ref{fig:related_work}(a). In the case of LayoutLMv1~\cite{xu2020layoutlm}, obtained text strings and coordinates are inputted into a BERT-based Transformer to acquire embeddings for downstream tasks. 

Furthermore, to enhance the performance, LayoutLMv2 and v3~\cite{xu2020layoutlmv2,huang2022layoutlmv3} also input image embeddings into the Transformer. DocFormer~\cite{Appalaraju_2021_ICCV} utilizes effective pre-training tasks designed for VDU tasks. BROS~\cite{hong2022bros} utilizes the relative positions of texts in 2D space. RDU~\cite{zhu2022rdu} employs a region prediction approach instead of sequence tagging. DocTR~\cite{Liao_2023_ICCV} represents text entities and links using anchor words inspired by anchor-based object detectors. VGT~\cite{Da_2023_ICCV} employs a grid transformer tailored for document layout analysis.

These approaches have shown promising performance, but they suffer from high computational costs, inflexibility on languages, and error propagation to the subsequent process due to using an off-the-shelf OCR.
To address these issues, VIES~\cite{wang2021towards} and CFAM~\cite{kuang2023visual} employ trainable text detectors like R-CNN~\cite{Girshick_2014_CVPR} instead of OCR models. As shown in Fig.~\ref{fig:related_work}(b), although these models consist of various modules such as a text detector, recognizer, and parser, they enable end-to-end optimization due to utilizing feature-level integration.

\subsection{OCR-Free Approach}
OCR based approaches that utilize various modules often suffer from the complexity of their model structures. 
To solve these problems, OCR-free end-to-end approaches using sequence generation have been proposed that simplify the model architecture and achieve cost-effectiveness, as shown in Fig.~\ref{fig:related_work}(c). Donut (Document Understanding Transformer)~\cite{kim2022donut} utilizes the transformer framework for the direct mapping of an input document image into the desired structured output. Additionally, StrucTextv2~\cite{yu2023structextv} proposes an effective pre-training framework using only image input and has enhanced performance in VDU tasks.

The approaches that indirectly extract text coordinates from attention maps~\cite{kim2023text} or from area of interest masks~\cite{Cao_2023_ICCV} indicated that text coordinate information could already be embedded into features. 
As a next step, while employing OCR-free end-to-end approaches, the proposed CREPE extracts text coordinates explicitly through a trainable coordinate head, as shown in Fig~\ref{fig:related_work}(d).

\section{Methods}

\subsection{Overview}

\begin{figure}[t]
\centering
\includegraphics[width=\textwidth]{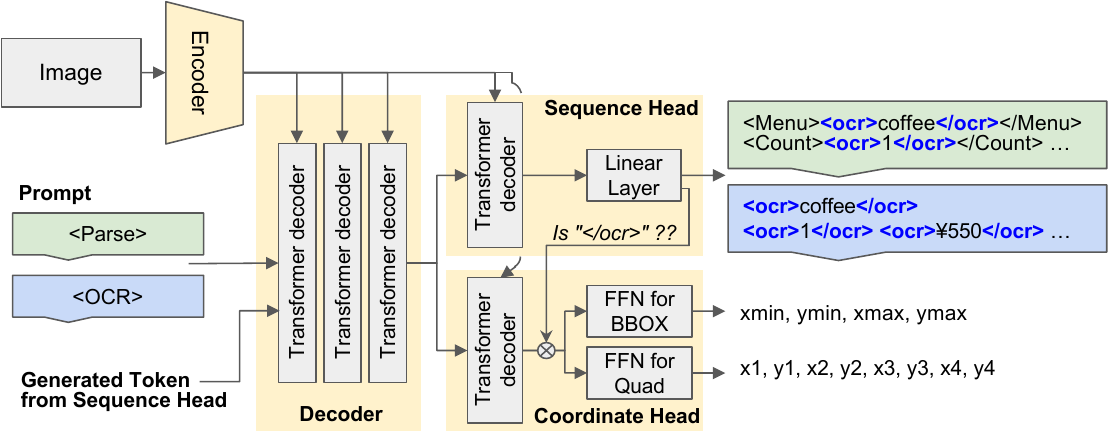}
\vspace{-6mm}
\caption{
Overview of CREPE Architecture. CREPE employs a multi-head architecture designed to generate parsing outputs and corresponding text coordinates concurrently. It utilizes the special token \texttt{</ocr>} that indicates a text segment is associated with coordinates output from different heads.
}
\label{overview}
\end{figure}

Fig.~\ref{overview} shows an overview of the proposed architecture. Similar to Donut~\cite{kim2022donut}, CREPE inputs document images into a Swin Transformer encoder~\cite{Liu_2021_ICCV}, then feeds the visual features along with task-specific prompts into a text decoder based on mBART~\cite{liu2020multilingual}. The model generates tokens via the text decoder and feeds the generated token back into the text decoder to predict the next token. 
In addition, CREPE adopts a multi-head architecture and simultaneously provide parsing outputs and text coordinates. These separated heads can be beneficial when training the model without coordinate ground truth. To appropriately associate text strings with coordinates output from different heads, we incorporates an output control mechanism using special OCR tokens, inspired by MTL-TabNet~\cite{visapp23namly} and UniTAB~\cite{yang2022unitab}.

\begin{figure}[t]
\includegraphics[width=\textwidth]{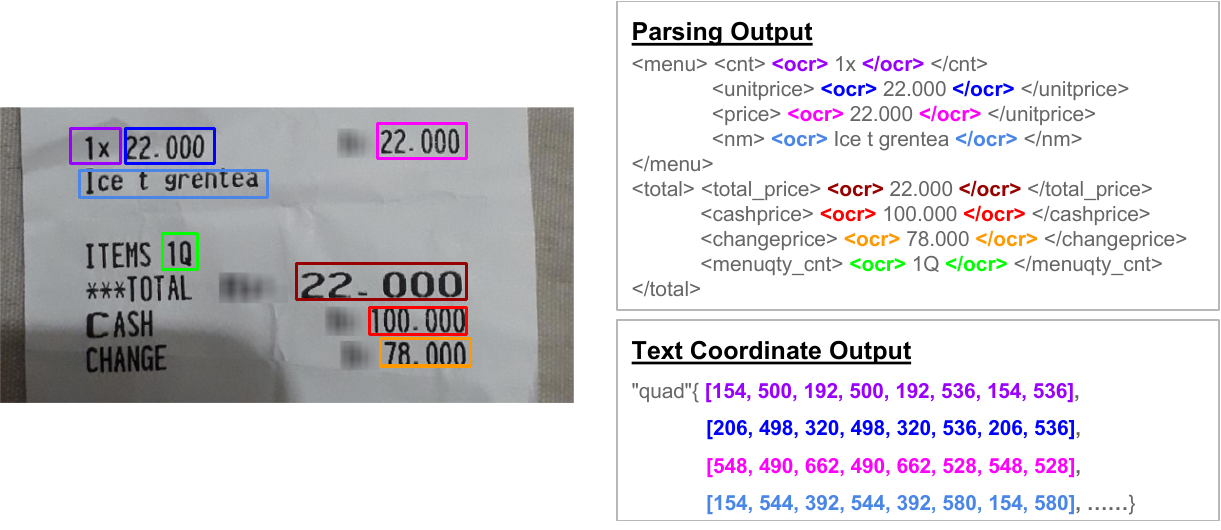}
\vspace{-7mm}
\caption{
Sample Output from CREPE. The converted output comprises both parsing results and associated text coordinates. Corresponding outputs are highlighted using matching colors. The alignment between the text and coordinates is facilitated through the use of the special token \texttt{</ocr>}.
} 
\label{fig:output_example}
\end{figure}

\subsection{Special tokens for OCR representation}
\label{Special_Token}
Donut~\cite{kim2022donut} converts the structured data in JSON format into a sequence of tokens by adding start and end special tokens for each semantics \texttt{<*>} and \texttt{</*>}. For example, if there is 'hot coffee' whose semantic is the menu name, the converted tokens become like below.

\vspace{2mm}
\fbox{\texttt{<menu>} \texttt{hot} \texttt{coffee} \texttt{</menu>}}
\vspace{2mm}

In our method, the special tokens for structure representation are identical to Donut's. In addition to that, we introduce special tokens for OCR texts. The tokens \texttt{<ocr>} and \texttt{</ocr>} denote the start and end of each text segment, playing a role in separating each text instance by enclosing text string within the generated sequence. In the case of 'hot coffee', The converted tokens become like the below.

\vspace{2mm}
\fbox{\texttt{<menu>}\texttt{<ocr>}\texttt{hot}\texttt{</ocr>}\texttt{<ocr>}\texttt{coffee}\texttt{</ocr>}\texttt{</menu>}}

\subsection{Multi-head architecture for text and coordinate}
We use four transformer layers for decoding text sequences. Among them, three bottom layers are shared, and an additional transformer decoder is attached to separate sequence and coordinate heads. The coordinate head can predict bounding boxes ($x_{min}$, $y_{min}$, $x_{max}$, $y_{max}$) as well as quadrilaterals ($x_1$, $y_1$, $x_2$, $y_2$, $x_3$, $y_3$, $x_4$, $y_4$). Two box-type branches of the coordinate head are composed of a three-layer feed-forward network (FFN) like~\cite{carion2020end}. These two types of outputs can be optionally utilized depending on the document type.

This coordinate head is the distinctive and unique feature of CREPE.
Alternatively, another approach could involve incorporating coordinate prediction into a sequence generation model to generate normalized coordinate tokens~\cite{liu2023spts,kil2023towards,chen2021pix2seq}. However, such an approach may result in outputs that are overly long and redundant; i.e., eight position tokens are necessary to represent the quadrilateral bounding box.
In addition, these methods would require coordinates to be provided, thus sacrificing the advantages of an end-to-end parsing model. The proposed multi-head architecture not only avoids the redundancy of output sequences but also enables cost-effective learning without text coordinate annotations. This will be further explained in the subsection \ref{Weakly_Supervised_Learning}.

Unlike the sequence head, which generates tokens at every step, the coordinate is triggered only when a specific token occurs, which is \texttt{</ocr>}, indicating the end of a specific text. The reason for using the \texttt{</ocr>} token as a trigger is that it contains the most accurate positional information since it represents the end of all decoded tokens in text instances. Using \texttt{</ocr>} as the trigger token showed more stable performance compared to using the start of OCR token \texttt{<ocr>}. The coordinates used in the coordinate head are normalized to the input size of the image encoder, ensuring values between 0 and 1. Examples of outputs generated in this way are depicted in Fig.~\ref{fig:output_example}.

To train the model, we compute cross-entropy loss and L1-loss for sequence head and coordinate head, respectively. 
Additionally, Distance Box IoU Loss~\cite{zheng2020distance} is computed for bounding boxes, which has been observed to expedite training speed.

\begin{figure}[t]
\includegraphics[width=\textwidth]{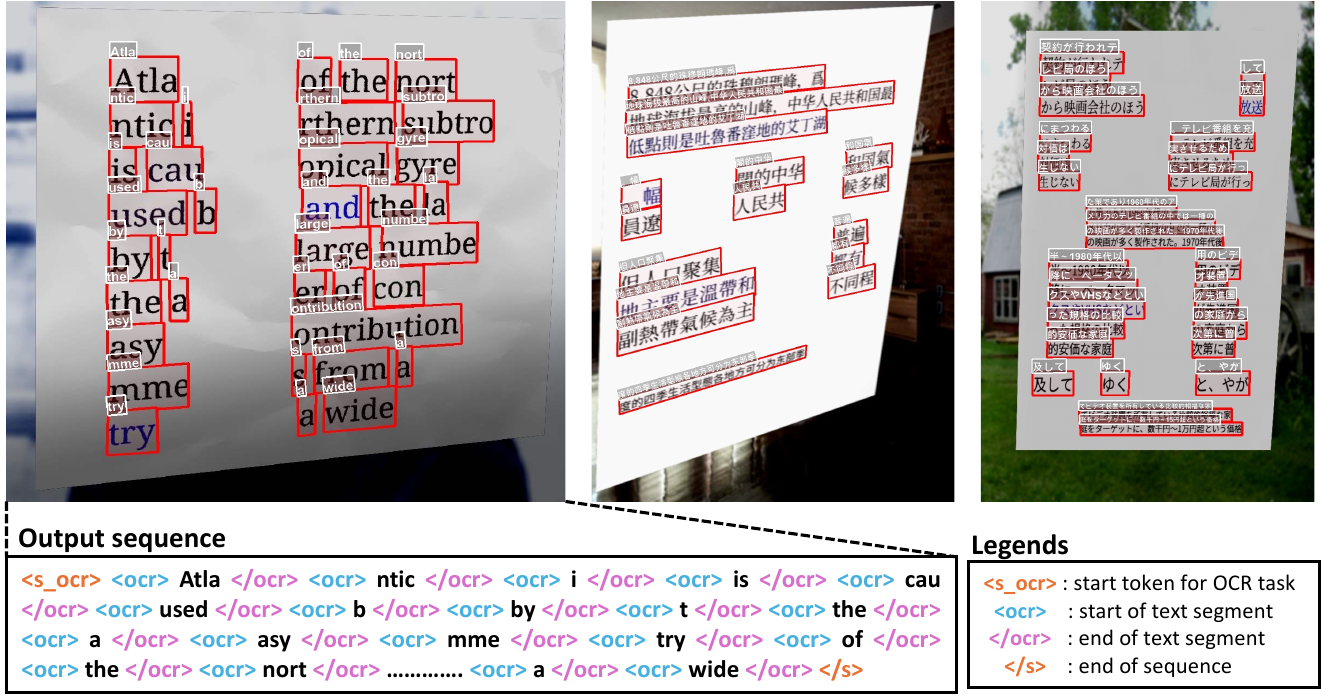}
\vspace{-6mm}
\caption{
OCR Task Result of the Pretrained Model. The text sequence and the corresponding bounding boxes were obtained from sequence head and coordinate head, respectively.}
\label{fig:result_pretraining}
\end{figure}

\subsection{Pre-training OCR functionality}
\label{pretraining_ocr}
The pre-training task of Donut~\cite{kim2022donut} was text reading, which is to generate all the text sequences in the document image. To obtain numerous visual corpora, synthetic data was generated using SynthDoG~\cite{yim2021synthtiger}. However, since the datasets generated in this way lack coordinate information of the text, we could not train the model including the coordinate head.

Therefore, we modified SynthDoG to generate text sequences and quadrilateral coordinates corresponding to each text instance. The amount of data is 50k samples per language, including English, Chinese, Japanese, and Korean. We trained the model using those visual corpora. The training resolution was 1280x1280, with a learning rate of 3e-5, a batch size of 4, and for 100k iterations. The task start token was specified as \texttt{<s\textunderscore ocr>}. When the OCR task token is given, we get the OCR results in Fig.~\ref{fig:result_pretraining}. As seen in the results, the pre-trained model is able to perform the basic OCR function of text detection and recognition in an end-to-end manner.

\subsection{Weakly Supervised Learning Framework}
\label{Weakly_Supervised_Learning}

\begin{figure}[t]
\includegraphics[width=\textwidth]{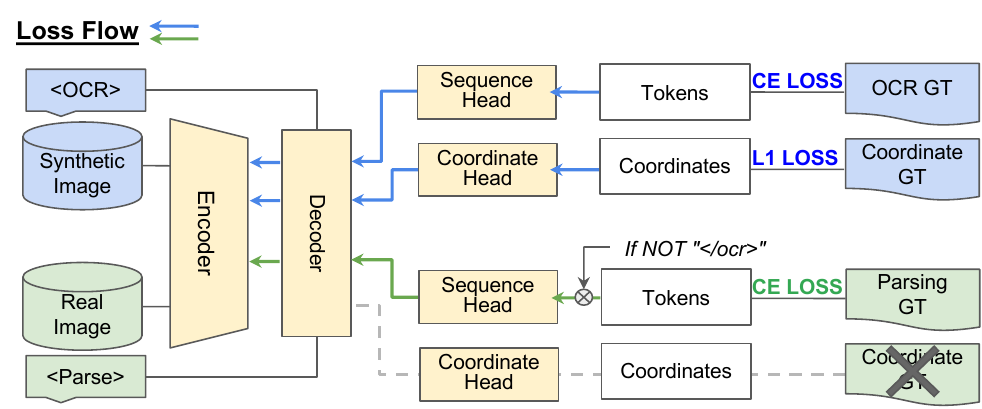}
\caption{
Weakly Supervised Learning Framework. The framework involves training on OCR tasks using synthesized document images and training on parsing tasks using real document images. 
It enables the CREPE model to be trained even in scenarios where text coordinate annotations are unavailable.}
\label{fig:Weakly_Supervised_Learning}
\end{figure}

After the model has learned OCR functionality, we fine-tune it to perform downstream tasks. For instance, we teach the model to extract key-value pairs for information extraction. Basically, training multi-heads requires text and coordinate pair data, which contradicts the advantage of sequence generation models that do not require text coordinates. To tackle this issue, we propose weakly-supervised training, which can be learned even without coordinate information.

Our proposed training scheme gets inputs from OCR datasets and parsing datasets in the same batch. Considering the risk that training the model without coordinate information makes the model forget its localization capability. Therefore, as illustrated in Fig.\ref{fig:Weakly_Supervised_Learning}, we adopt two techniques to maintain the behavior of \texttt{</ocr>}. Firstly, we mix synthetic and real data in a single batch during training. This allows the model to learn both the OCR task and the parsing task together, where the OCR task provides positional information for \texttt{</ocr>} by calculating coordinate loss. Secondly, during training the parsing task, we ignore the loss calculation for \texttt{</ocr>} token in the teach forcing scheme. By controlling the loss in the sequence, we can achieve the localization capability of \texttt{</ocr>} even in the parsing task.

\section{Experiments}
\label{Experiments}

\subsection{Parsing Performance Evaluation}
\label{VDU_Results}

To demonstrate the effectiveness of CREPE, we evaluated its performance in parsing tasks using publicly available datasets.
For CORD~\cite{park2019cord}, POIE~\cite{kuang2023visual}, and FUNSD~\cite{jaume2019funsd} datasets which include both parsing and text coordinate annotations, we conducted supervised learning for downstream tasks on each of the datasets. 
In contrast, for the TrainTicket~\cite{guo2019eaten} dataset, which lacks text coordinate annotations, we employed the proposed weakly-supervised learning framework. We composed a single batch for parsing and OCR tasks in a 1:4 ratio.

\begin{table}[t]
  \centering
  \caption{
  Parsing Performance Evaluations.
  The OCR-dependent methods require text input, thus during performance evaluation, they use the ground truth of text strings and coordinates as model inputs instead of OCR outputs.
  Therefore, directly comparing OCR-dependent models with generative models like ours is not feasible.
  On the FUNSD dataset, the evaluation metrics used were inconsistent across methods. \cite{hong2022bros,huang2022layoutlmv3,Liao_2023_ICCV} used the F1 score, while ours and \cite{yu2023structextv} employed the Tree Edit Distance.
  $\dagger$ indicates the number of params without multi-lingual vocabulary for fair comparison, and $\alpha$ means additional params for specific task.}
  \vspace{2mm}
  \label{tab:vdu_results}
  \begin{tabular}{l|c|c||c|c|c|c}
    \hline
    \textbf{Model} & \textbf{Params} & \textbf{OCR} & \makecell{\textbf{CORD}\\\textit{(F1)}} & \makecell{\textbf{TrainTicket}\\\textit{(F1)}} & \makecell{\textbf{POIE}\\\textit{(F1)}} & \makecell{\textbf{FUNSD}\\\textit{(F1/TED)}} \\ 
    \Xhline{2\arrayrulewidth}
    BROS~\cite{hong2022bros}   & $340_M$ &\checkmark& 97.4 & & & 84.5 \\
    DocFormer~\cite{Appalaraju_2021_ICCV}      &$183_M$ &\checkmark&      96.3 &          &           & 83.3 \\
    LayoutLMv3~\cite{huang2022layoutlmv3}     &$368_M$ &\checkmark&      97.5&          &           & 92.1 \\
    TILT~\cite{powalski2021going} & $780_M$ &\checkmark&  96.3 & & & \\
    DocTr~\cite{Liao_2023_ICCV}          &$153_M$ &\checkmark&      94.4 &          &           & 73.9 \\ 
    \Xhline{2\arrayrulewidth}
    Donut~\cite{kim2022donut}          &$143_M^\dagger$ &&      84.1 &     94.1  &           & \\
    SeRum\textsubscript{total}~\cite{Cao_2023_ICCV}          &$136_M^\dagger$ &&      80.5 &     97.9  &           & \\
    TRIE~\cite{Zhang2020TRIEET}           &- &&           &           & 76.4      & \\
    VIES~\cite{wang2021towards}           &- &&           &           & 77.2      & \\
    CFAM~\cite{kuang2023visual}           &- &&           &           & 79.5  & \\
    StrucTextv2\textsubscript{s}~\cite{yu2023structextv}    &$28_M^\dagger+\alpha$ &&           &           &           & \textbf{55.0} \\ 
    \hline
    \textbf{CREPE (our)}   & $160_M^\dagger$ && \textbf{85.0} & \textbf{98.4}  & \textbf{79.6}  & 53.8 \\ 
    \hline
  \end{tabular}
\end{table}

The evaluation results are presented in Table~\ref{tab:vdu_results}.
In CORD, TrainTicket, and PIOE datasets, our CREPE achieved state-of-the-art performance within the same modality.

This suggests that a multi-head design would enhance parsing performance by facilitating complementary learning.
As an additional note, the performance difference between Donut~\cite{kim2022donut} and CREPE on TrainTicket is notably larger, despite being almost comparable on CORD, likely due to Donut's insufficient vocabulary.

With the addition of the coordinate head in the Donut model architecture, the parameter count increased by 17M. However, there is not a significant difference in inference speed since batch processing for coordinate prediction is performed by collecting the \texttt{</ocr>} tokens after sequence generation.

\begin{figure}[p]
\centering
\begin{minipage}{\textwidth}
\centering
\begin{tabular}{|m{0.12\textwidth}|m{0.29\textwidth}|m{0.29\textwidth}|m{0.29\textwidth}|}
\hline
\textbf{CORD} 
     & \includegraphics[width=\linewidth]{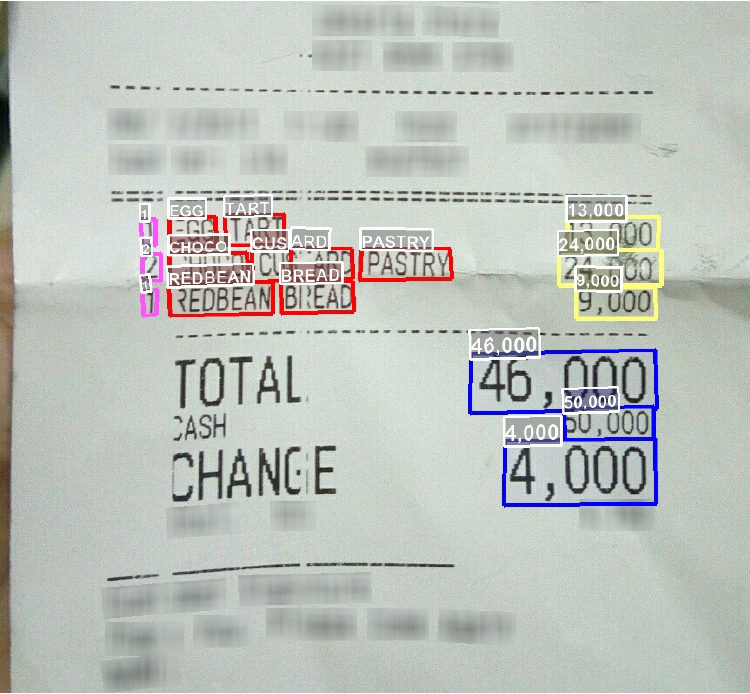} 
     & \includegraphics[width=\linewidth]{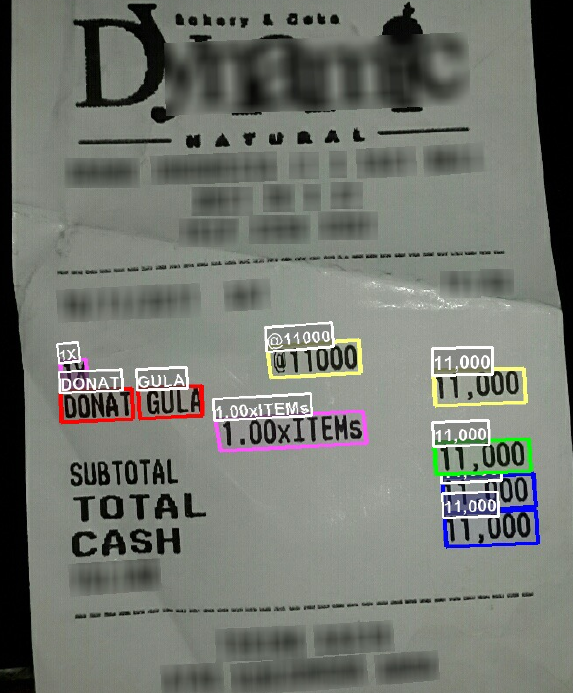} 
     & \includegraphics[width=\linewidth]{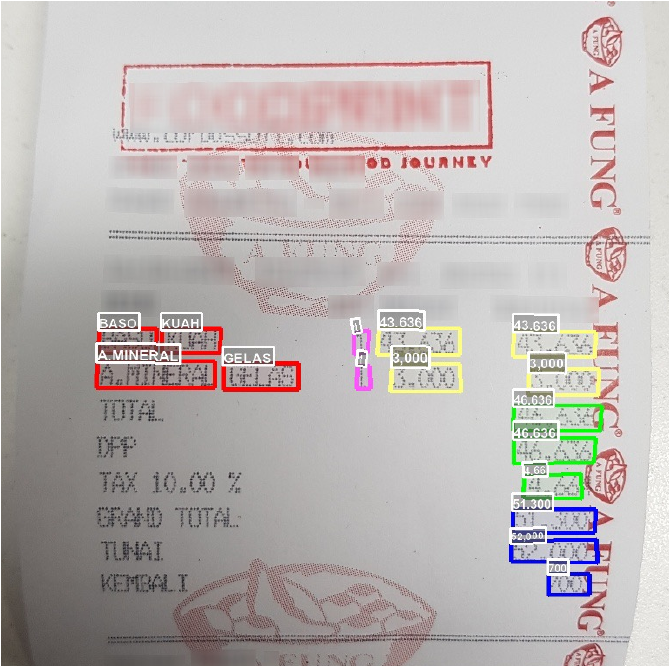} \\
\hline
\textbf{POIE}
     & \includegraphics[width=\linewidth]{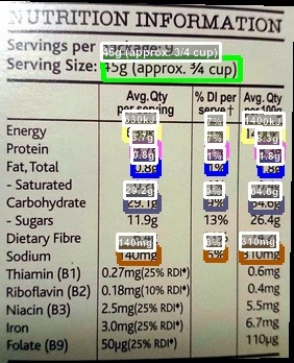} 
     & \includegraphics[width=\linewidth]{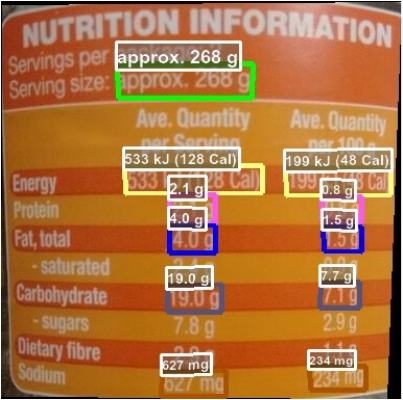} 
     & \includegraphics[width=\linewidth]{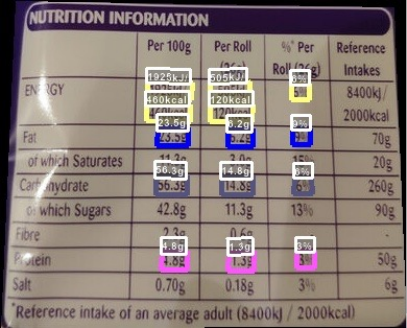} \\
\hline
\textbf{FUNSD}
      & \includegraphics[width=\linewidth]{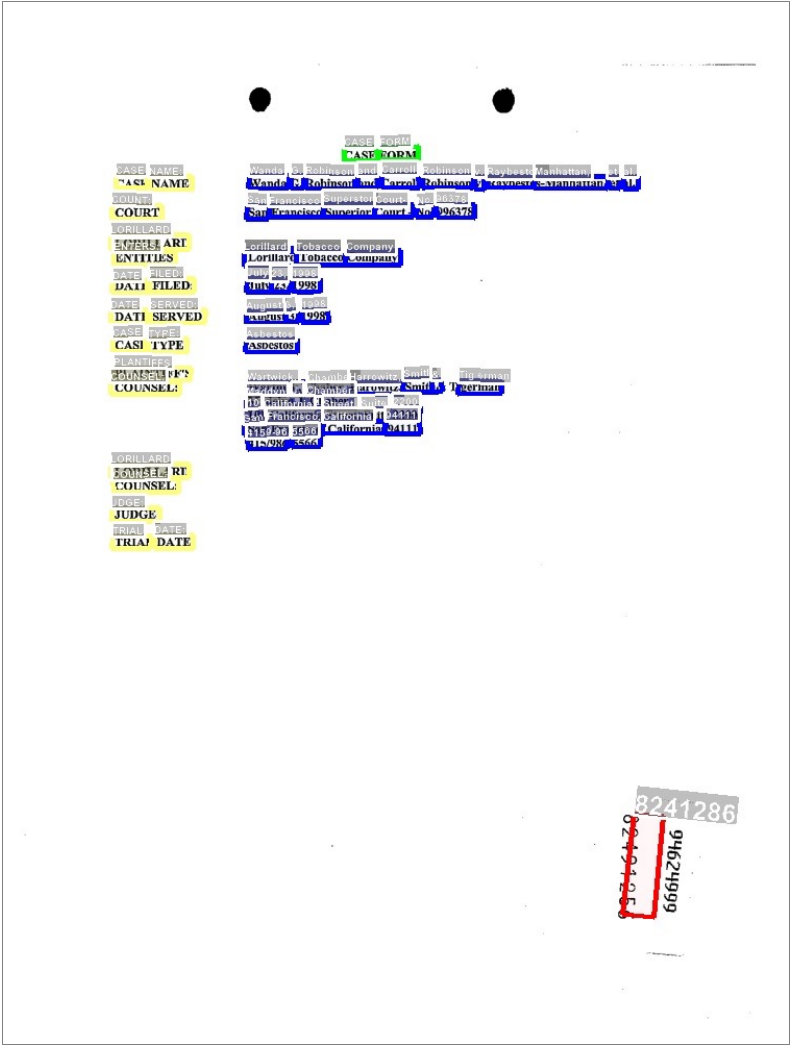} 
      & \includegraphics[width=\linewidth]{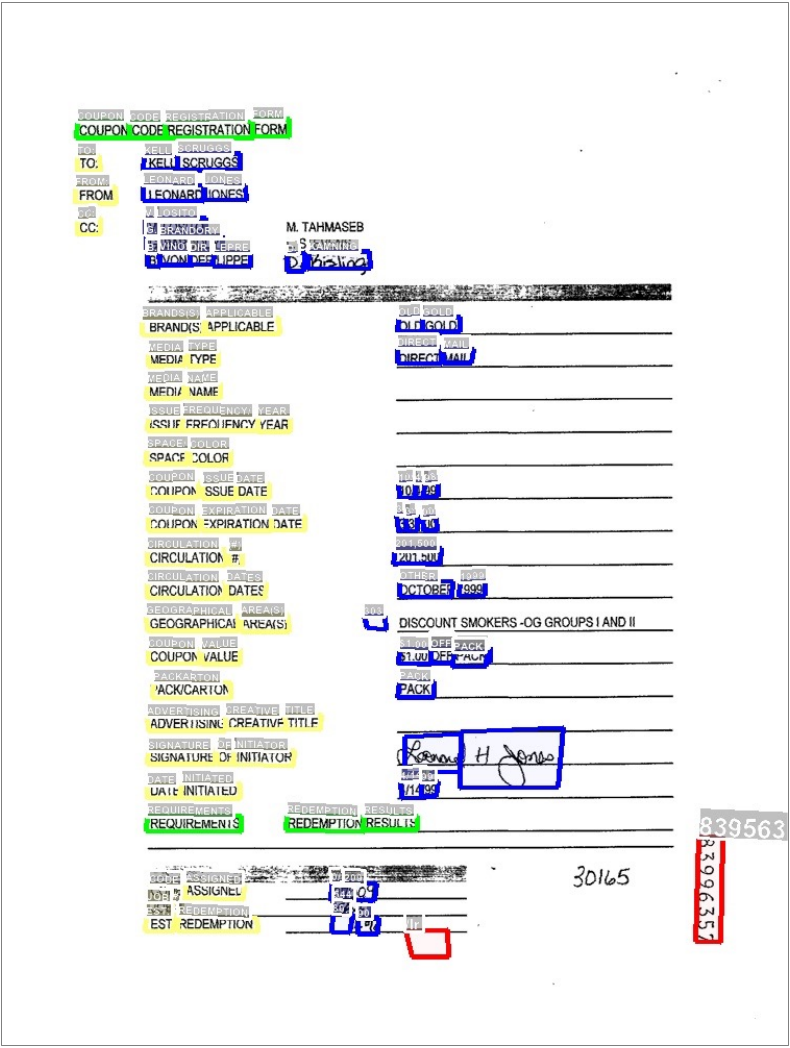} 
      & \includegraphics[width=\linewidth]{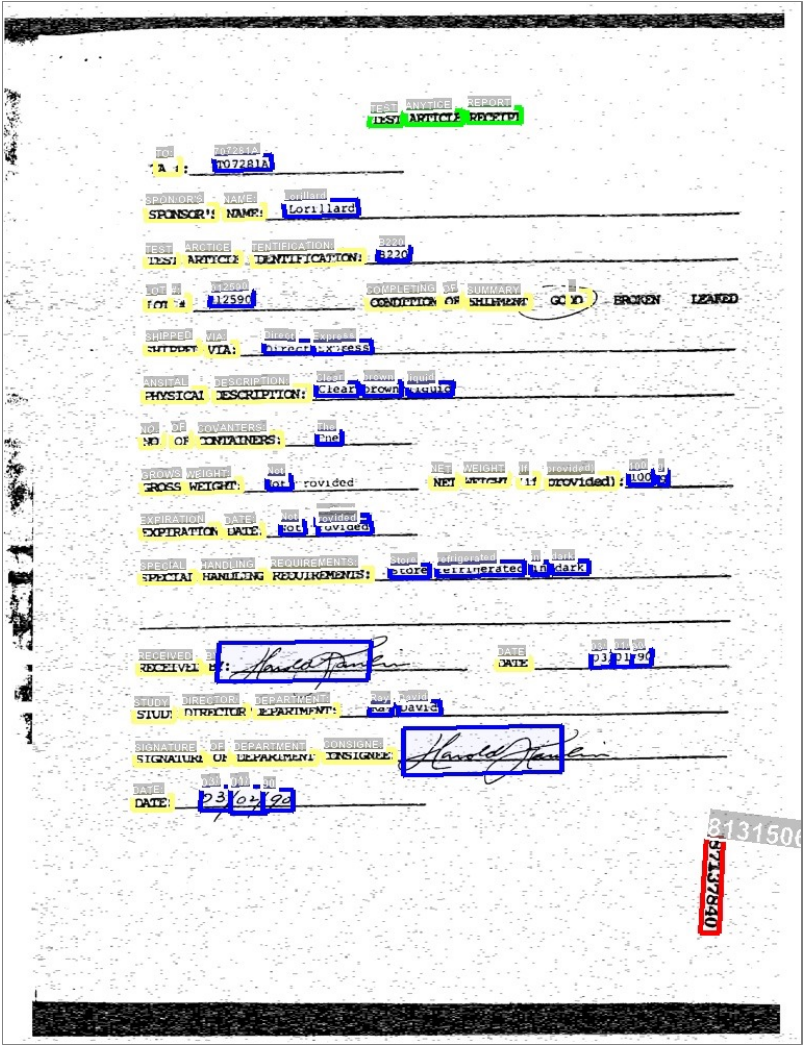} \\
\hline
\end{tabular}
The output of the coordinate head trained with supervised learning.
\end{minipage}

\vspace{4mm}

\begin{minipage}{\textwidth}
\centering
\begin{tabular}{|m{0.12\textwidth}|m{0.29\textwidth}|m{0.29\textwidth}|m{0.29\textwidth}|}
\hline
\textbf{Train Ticket} &
\includegraphics[width=\linewidth]{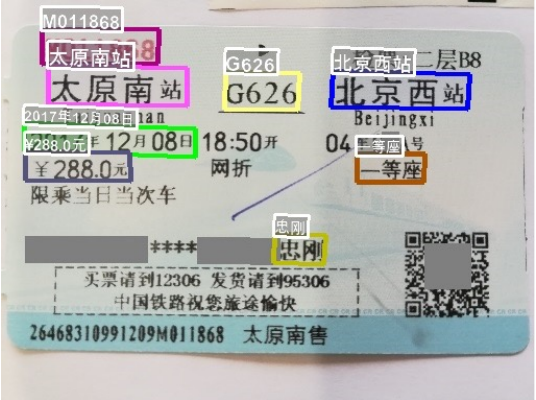} & \includegraphics[width=\linewidth]{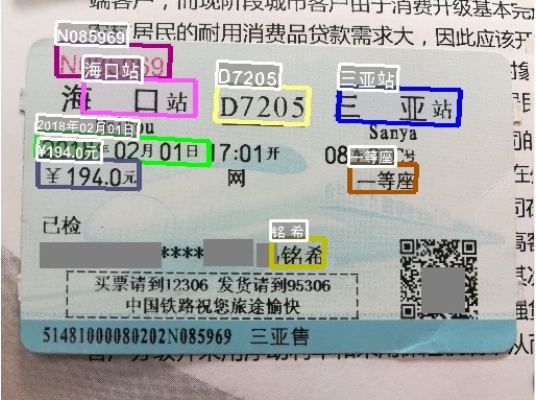} & \includegraphics[width=\linewidth]{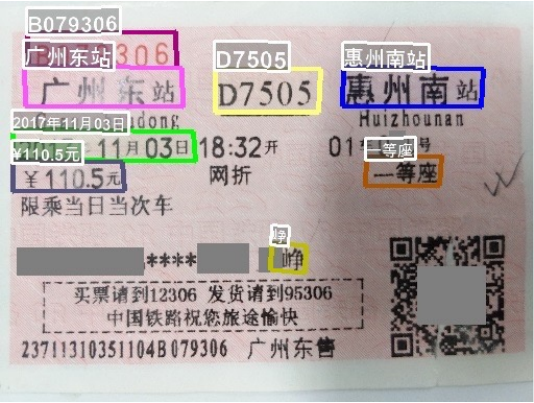} \\
\hline
\end{tabular}
The output of the coordinate head trained without supervision.
\end{minipage}
\caption{
The Results of Acquiring Text Coordinates.
CREPE is capable of acquiring coordinates for text strings contained within the parsing output.
}
\label{fig:Coordinate_Results}
\end{figure}

\subsection{Text Coordinate Results}
\label{Coordinate_Results}
Fig.~\ref{fig:Coordinate_Results} illustrates the results of text coordinate acquisition by CREPE. The color represents the provided semantics of the text for each domain. The results highlight that the proposed method successfully provides text coordinates.
To quantitatively assess, we utilized CLEval~\cite{baek2020cleval}, and Table~\ref{tab:coordinate_performance_evaluation} shows the evaluation results.
Unlike traditional evaluations on OCR tasks that recognize all text in images, our method only extract the value texts relevant to information extraction.
Therefore, in this evaluation, all text other than this key information was considered as 'don't care' for assessment.

It is difficult to claim that the score alone is sufficiently accurate; however, comparing evaluations is challenging due to the differing nature of the extracted text compared to conventional methods. In this paper, we have preserved it as a baseline for future comparisons with other research.

\vspace{-4mm}
\begin{table}[h]
\centering
\caption{Evaluation of text localization performance via CLEval}
\label{tab:coordinate_performance_evaluation}
\setlength\tabcolsep{12pt} 
\begin{tabular}{lccc}
\toprule
      & Recall & Precision & F1   \\
\midrule
CORD  & 95.1   & 95.8      & 95.5 \\
POIE  & 86.1   & 92.9      & 89.4 \\
\bottomrule
\end{tabular}
\end{table}

For the TrainTicket dataset, the ground truth only provides 8 field key-value pairs including the ticket number, starting station, train number, and so on. Our model was trained on only key-value ground truth without positional information for the characters, yet it was able to extract positional information as demonstrated in Fig.~\ref{fig:Coordinate_Results}. This verified the feasibility of the weakly supervised training framework, as described in Section~\ref{Weakly_Supervised_Learning}.

\newpage
\vspace{-4mm}
\subsection{Ablation Study}
\label{Ablation_Study}
To further validate the proposed weakly-supervised learning framework, we trained CREPE models under both supervised and weakly-supervised learning manners, while ensuring consistency in dataset and hyper parameters.

\begin{figure}[htp]
\centering
\begin{tabular}{|c|c|}
\hline
\textbf{Supervised Learning} & \textbf{Weakly Supervised Learning} 
\\
\hline
\parbox[c][4.5cm][c]{0.50\textwidth}{\centering\includegraphics[width=0.95\linewidth]{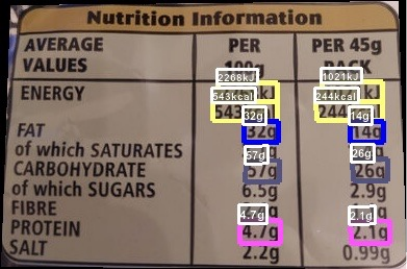}} &
\parbox[c][4.5cm][c]{0.50\textwidth}{\centering\includegraphics[width=0.95\linewidth]{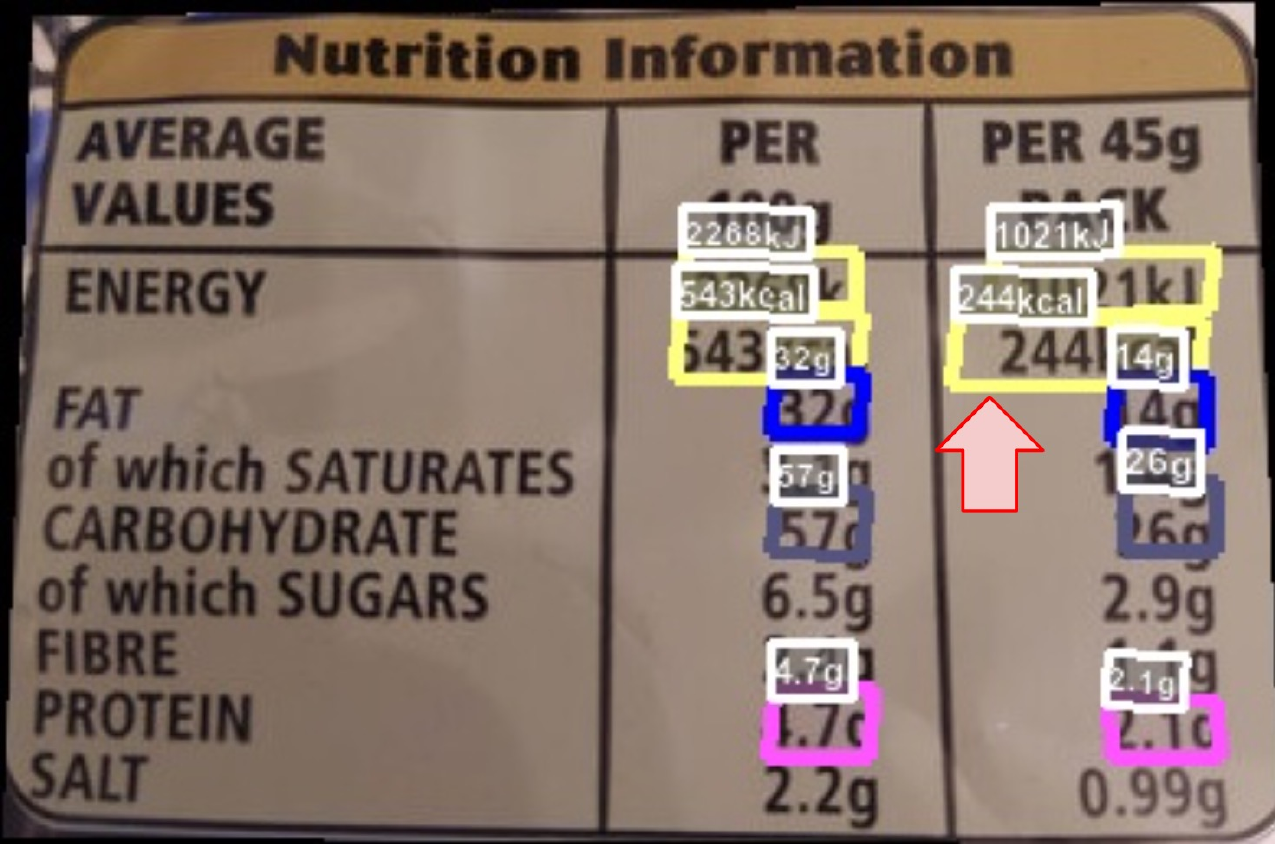}} 
\\
\hline
\parbox[c][6.5cm][c]{0.50\textwidth}{\centering\includegraphics[width=0.95\linewidth]{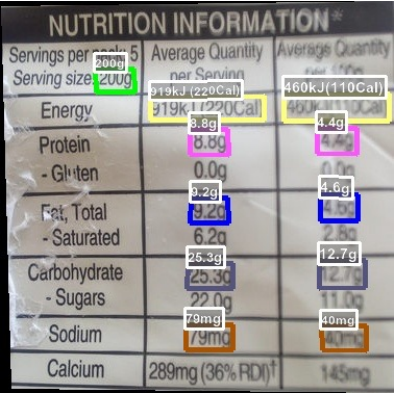}} &
\parbox[c][6.5cm][c]{0.50\textwidth}{\centering\includegraphics[width=0.95\linewidth]{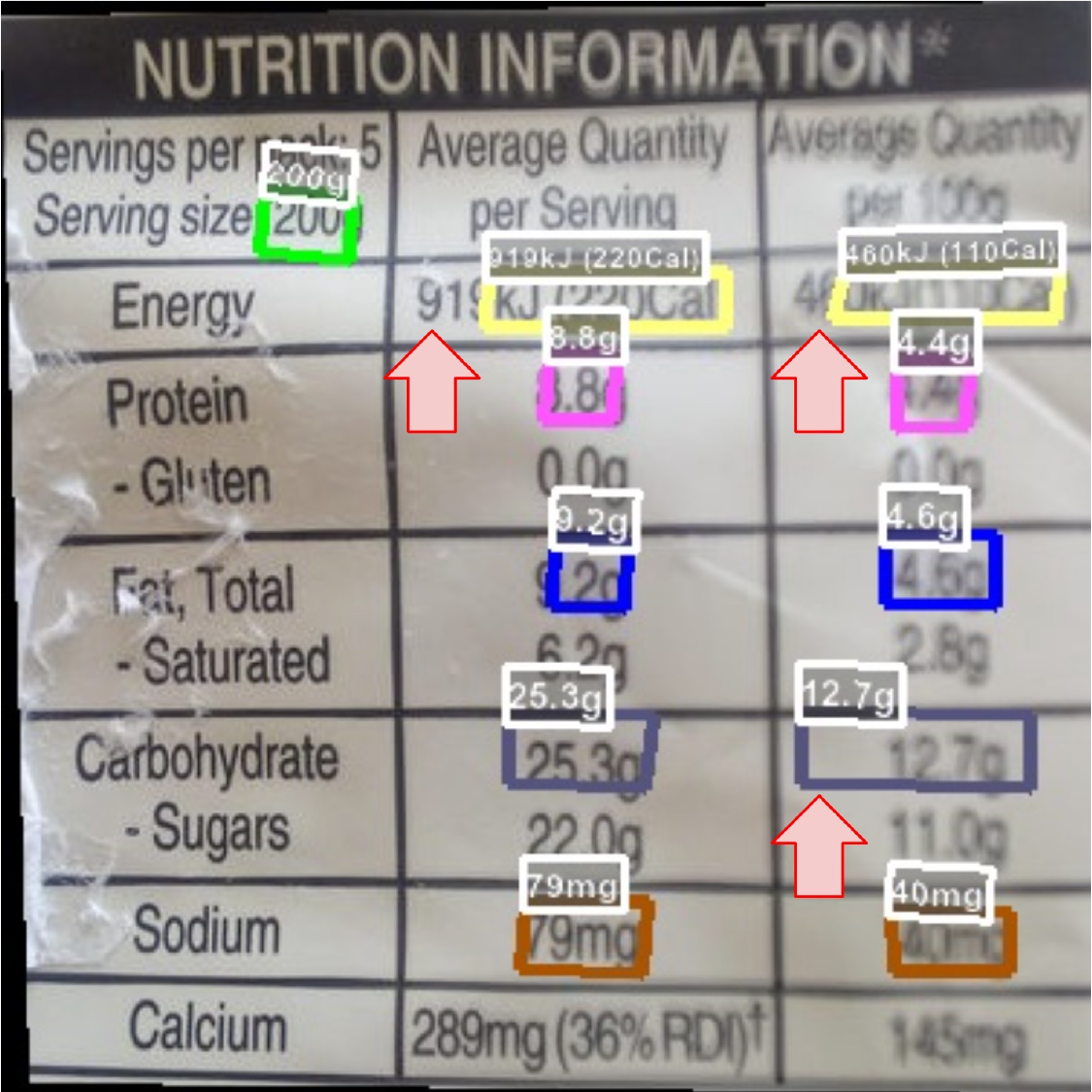}} 
\\
\hline
\end{tabular}
\caption{
Comparison of Cases of Supervised and Weakly-Supervised Learning.
Both approaches successfully capture the center points of the text. However, there was a difference in how accurately the model extracted the boundaries of the text boxes.
}
\label{fig:ablation_study}
\end{figure}

Fig.~\ref{fig:ablation_study} shows that some of the output bounding boxes exhibited noticeable distortion, when trained without explicit positional supervision. 
In particular, the text quadrilaterals failed to fully encompass certain portions of the text areas and unnecessarily extended into areas devoid of text. Nonetheless, even in these instances, the bounding boxes effectively captured the center point of the text. Therefore, the ongoing challenge lies in refining the quadrilaterals to accurately align with the boundaries of the text regions.

\newpage
\subsection{Adapting to Various VDU Tasks}
\label{Layout_Analysis}
This section demonstrates the adaptability of CREPE to various tasks by exploiting its localization capability of the designated sequence tokens.

\subsubsection{Document Image Classification Task}
We trained the sequence head to generate a special token indicating the document type label (e.g., \texttt{<letter>}, \texttt{<invoice>}) on the RVL-CDIP dataset~\cite{harley2015icdar}. Our CREPE model achieved an accuracy of 0.95, closely approaching the state-of-the-art performance of 0.97~\cite{bakkali2021eaml}.

\subsubsection{Layout Analysis Task} 
We trained the CREPE model on PubLayNet~\cite{zhong2019publaynet} dataset so that the sequence head generates a series of special tokens for indicating the layout labels and start/end of the layout, as described in Section~\ref{Special_Token}. In this case, \texttt{</layout>} is the trigger token for coordinate prediction. The converted tokens become like below.

\vspace{2mm}
\fbox{\texttt{<layout>} \texttt{<2\_title>} \texttt{</layout>} \texttt{<layout>} \texttt{<1\_text>} \texttt{</layout>} ...}
\vspace{2mm}

\begin{figure}[h]
\centering
\begin{tabular}{cc}
\includegraphics[width=0.5\textwidth]{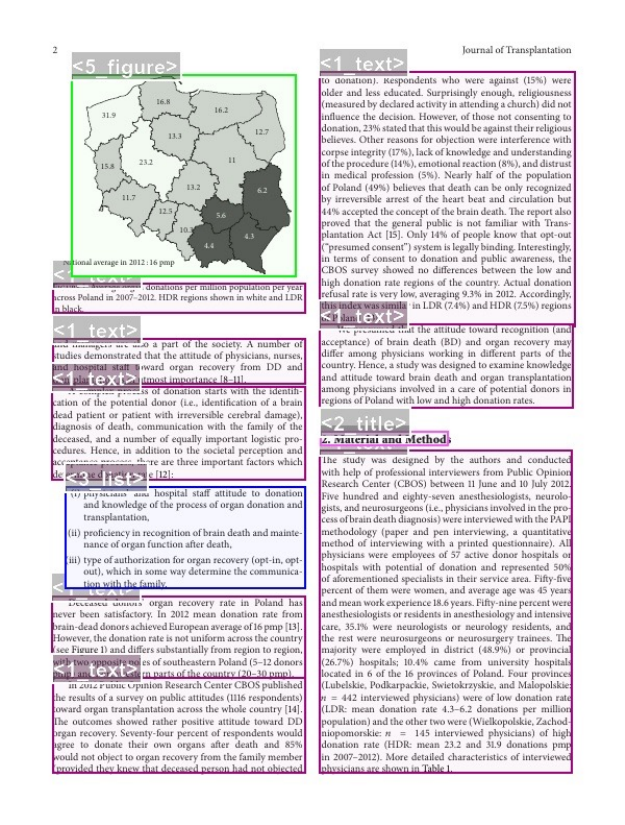} & 
\includegraphics[width=0.5\textwidth]{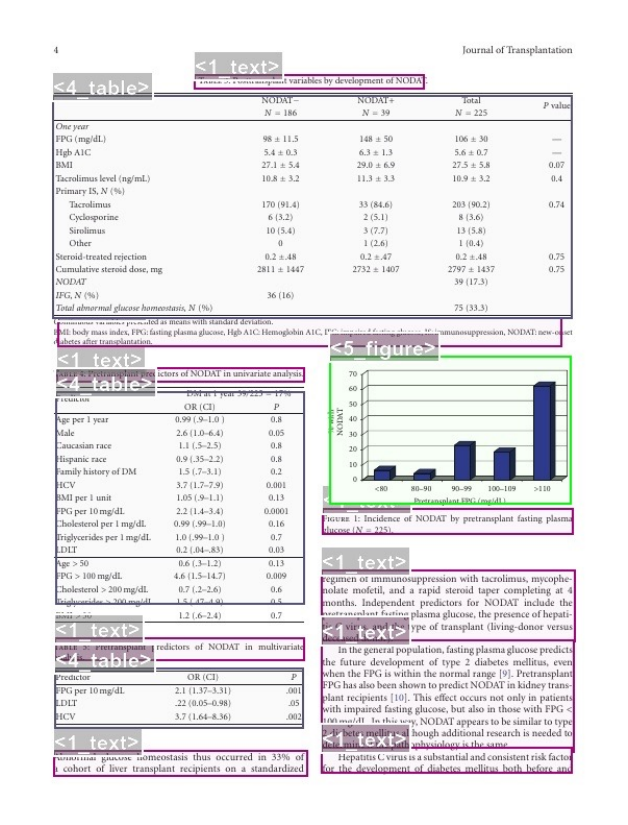} \\
\end{tabular}
\vspace{-6mm}
\caption{
The Result of the Layout Analysis Task.
CREPE enables inferring the layout and coordinates of each element within document images.
}
\label{fig:crepe_layout_analysis}
\end{figure}

As depicted in Fig.~\ref{fig:crepe_layout_analysis}, CREPE can infer the layout and coordinates of each element within a document. In quantitative evaluation, the mIoU was 0.658, leaving room for improvement. This limitation was attributed to the absence of clear rules for the order of layout elements, unlike cases such as the raster order for text elements, which occasionally resulted in missing elements.
The introduction of techniques such as sequence augmentation~\cite{chen2021pix2seq} would be required.

\vspace{-8mm}

\subsubsection{Document Visual Question Answering Task} 
\begin{figure}[ht]
\centering
\includegraphics[width=0.90\textwidth]{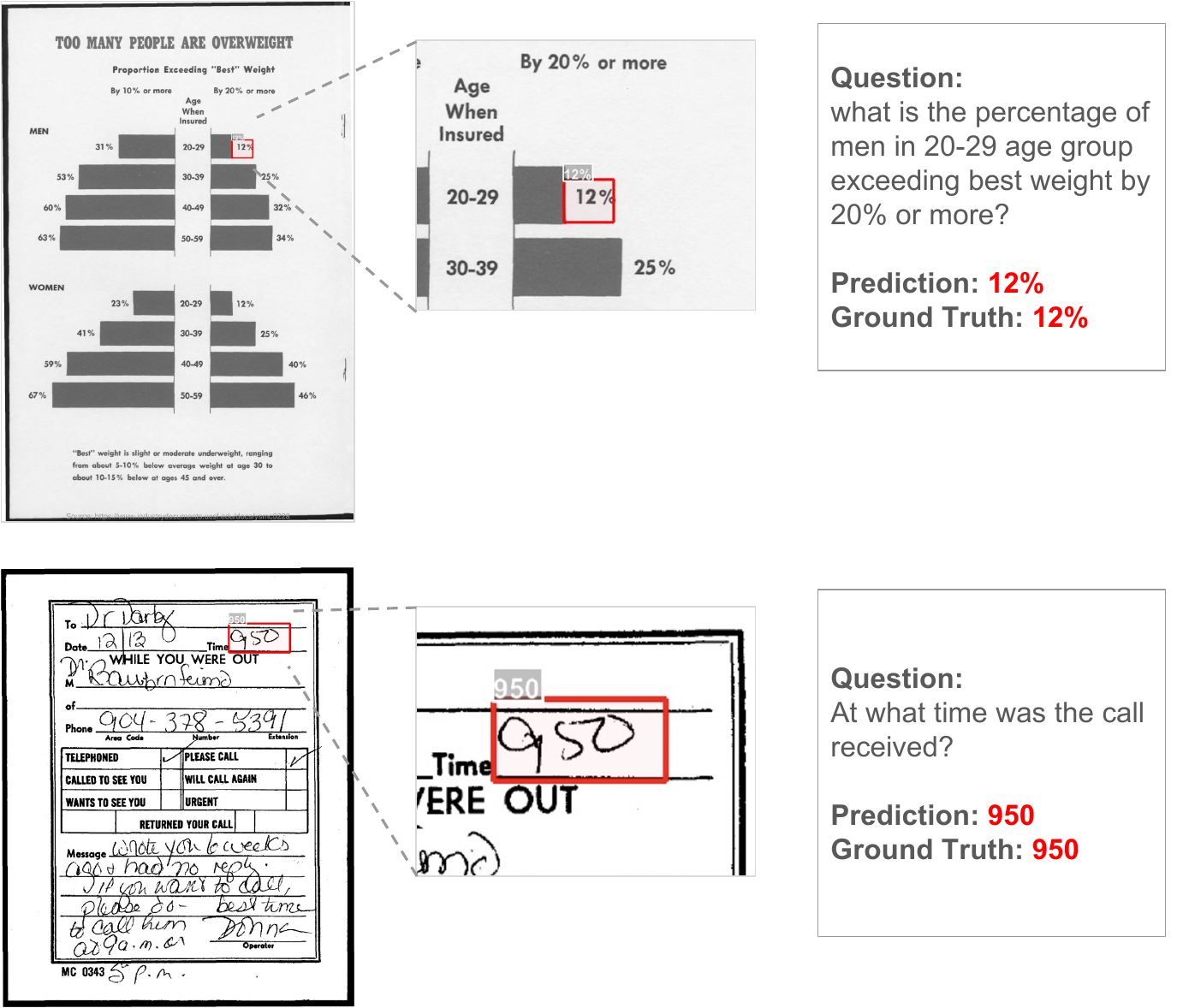}
\vspace{0mm}
\caption{
The Result of Document Visual Question Answering Task.
CREPE can provide contextual answers to questions and also indicate where the answer is located in the document image.}
\label{fig:VQA}
\end{figure}

The model was trained to generate an answer sentence for a question from a given document image using the DocVQA dataset~\cite{mathew2021docvqa}. Since the ground truth does not provide positional information for answer texts, we employed weakly-supervised learning in this experiment. As illustrated in Fig.~\ref{fig:VQA}, CREPE has demonstrated its capability to answer questions along with locating the answer texts within the document image.
Nonetheless, in quantitative evaluation, the ANLS stood at 58.36, indicating room for improvement in future work.

\vspace{-4mm}

\vspace{-6mm}
\subsection{CREPE in Scene Understanding: Beyond VDU Tasks}
As further investigation, we tackled the tasks of scene text detection~\cite{zhu2016scene} and object detection~\cite{zou2023object}, inspired by SPTSv2~\cite{liu2023spts} and Pix2Seq~\cite{chen2021pix2seq}.
Unlike VDU tasks, these tasks target scene images and involve identifying the categories and coordinates of objects within the images.
Traditionally, these detection tasks have been conducted separately. Yet, CREPE introduces a novel approach to perform them simultaneously, and it would enrich information acquisition in some domains, such as robot vision for autonomous driving or self-navigating.

We trained and evaluated our model on the COCO 2014 dataset~\cite{lin2014microsoft}, including scene images, object categories, coordinates, and text annotations from COCO Text v2~\cite{veit2016coco}.
To detect scene texts, we adopted the approach similar to that used in OCR tasks. To detect objects, we introduced new special tokens. First; \texttt{<obj>} and \texttt{</obj>} that denote the start and end of an object instance, respectively. \texttt{</obj>} is also used to align the detected object with its coordinates. Second; special tokens for each object category, allowing the model to describe object categories within the output text sequence. The converted tokens become like below.

\vspace{2mm}
\fbox{\texttt{<obj>}\texttt{<6\_bus>}\texttt{</obj>} \texttt{<ocr>}\texttt{201}\texttt{</ocr>}  \texttt{<ocr>}\texttt{Hyakumanben}\texttt{</ocr>}}
\vspace{2mm}

\begin{figure}[h]
\centering
\includegraphics[width=\textwidth]{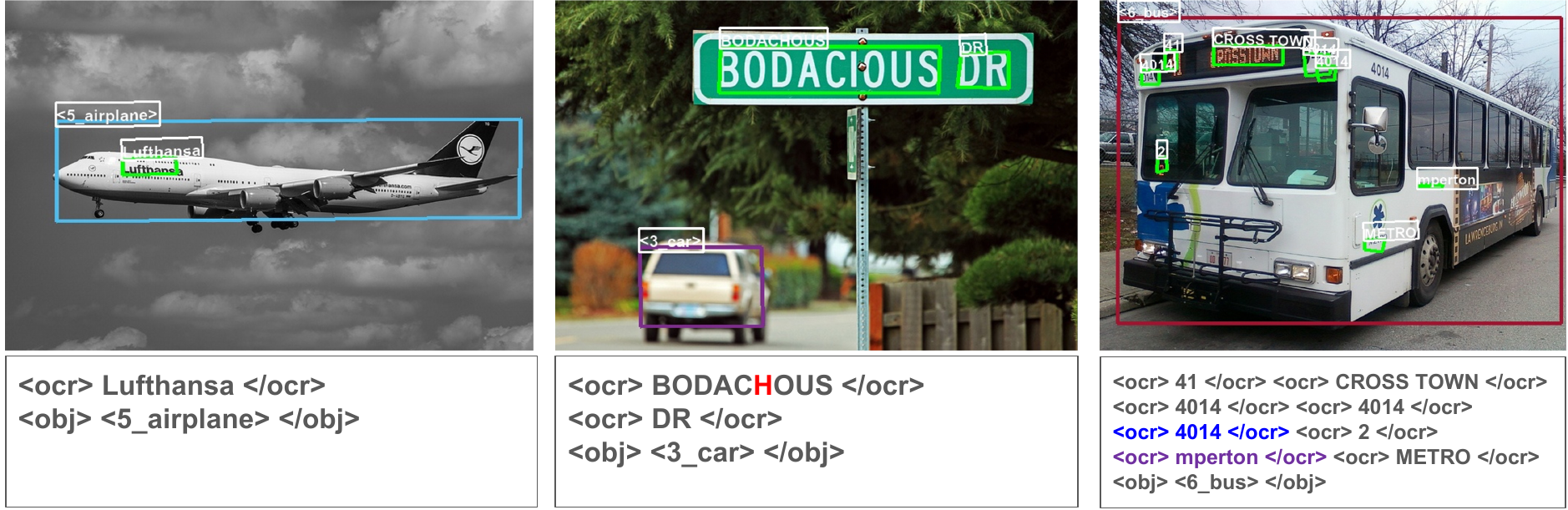}
\caption{
The Scene Text and Object Detection Results.
The text sequence highlighted in red indicates incorrect strings, blue indicates incorrect coordinates, and purple denotes the text that could not be deciphered from the image.
}
\label{fig:object_detection_example}
\end{figure}

Fig.~\ref{fig:object_detection_example} illustrates the results and demonstrates CREPE's capability of detecting scene texts and objects simultaneously.
However, similar to the limitations mentioned in pix2Seq\cite{chen2021pix2seq}, detection via the autoregressive approach tends to terminate before all objects are predicted, especially when attempting to detect a multitude of texts and objects, a limitation that applies to CREPE as well. Developing effective training methods to ensure accurate inference in such scenarios remains an imperative task for future research.

\section{Conclusion}
In this research, we introduced CREPE, a novel end-to-end architecture for VDU tasks capable of simultaneously providing parsing output and text coordinates. Our experiments demonstrated that CREPE achieved state-of-the-art performance in parsing across various datasets while extracting text coordinates. Additionally, we showed that CREPE could be trained even without text coordinate annotations by utilizing our proposed weakly supervised learning framework. As demonstrated with tasks like layout analysis, a future challenge is to extend its application to a broader range of document understanding tasks.

%
%

\bibliographystyle{splncs04}
\bibliography{ms}

\end{document}


\maketitle              

\vspace{-4mm}
\section{Multi-Document Parsing} 
\subsection{Motivation} 
\begin{figure}[h]
\centering
\includegraphics[width=0.9\textwidth]{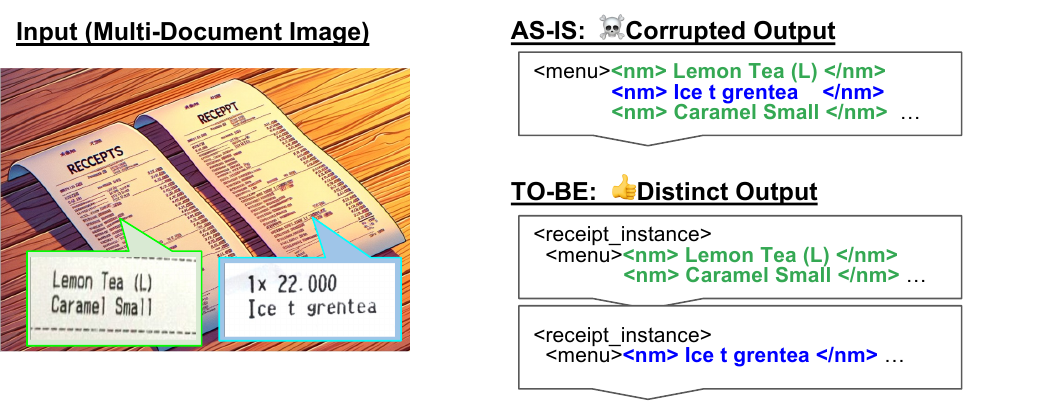}
\caption{
\textbf{The Remaining Challenge in Parsing Images Containing Multi-Document.} 
Traditional parsing models have implicitly assumed that each input image contains only a single document. Consequently, when an input image includes multiple documents, the resulting analysis may be incorrect, or the parsing outcomes for the two documents may become intertwined, leading to corrupted results.
}
\label{fig:issue}
\end{figure}

In this section, we explore the capability of proposed CREPE to provide the parsing results for each document in a distinguishable manner when given an image containing two or more documents.
Regardless of whether it is an OCR-based approach~\cite{huang2022layoutlmv3,Liao_2023_ICCV}  or OCR-Free approach~\cite{kim2022donut,Cao_2023_ICCV}, traditional parsing methods often implicitly assume that the input image contains only a single document; therefore, when presented with a multi-document image, they tend to produce corrupted output, as illustrated in Fig.~\ref{fig:issue}.
In scenarios where parsing models are used to enhance the efficiency of document processing tasks, especially when dealing with a large volume of documents, being restricted to inputting only one document at a time can inconvenience users and reduce computational resource efficiency.
To tackle this, we conduct a feasibility study on whether our proposed CREPE can simultaneously parse multiple documents in bulk.

\subsection{Approach} 
We train the CREPE model to generate a text sequence with special tokens that allow for the description of parsing results of multiple documents within a single sequence. Specifically, special tokens \texttt{<instance>} and \texttt{</instance>} are inserted in the sequence to enclose each parsing annotation segment, to ensure they are distinguishable even in a text sequence format.
Unlike using the document area detector~\cite{tensmeyer2017pagenet} to divide each document area, our approach allows for the bulk processing of multiple documents in a single inference. 

\begin{figure}[t]
\centering
\includegraphics[width=0.9\textwidth]{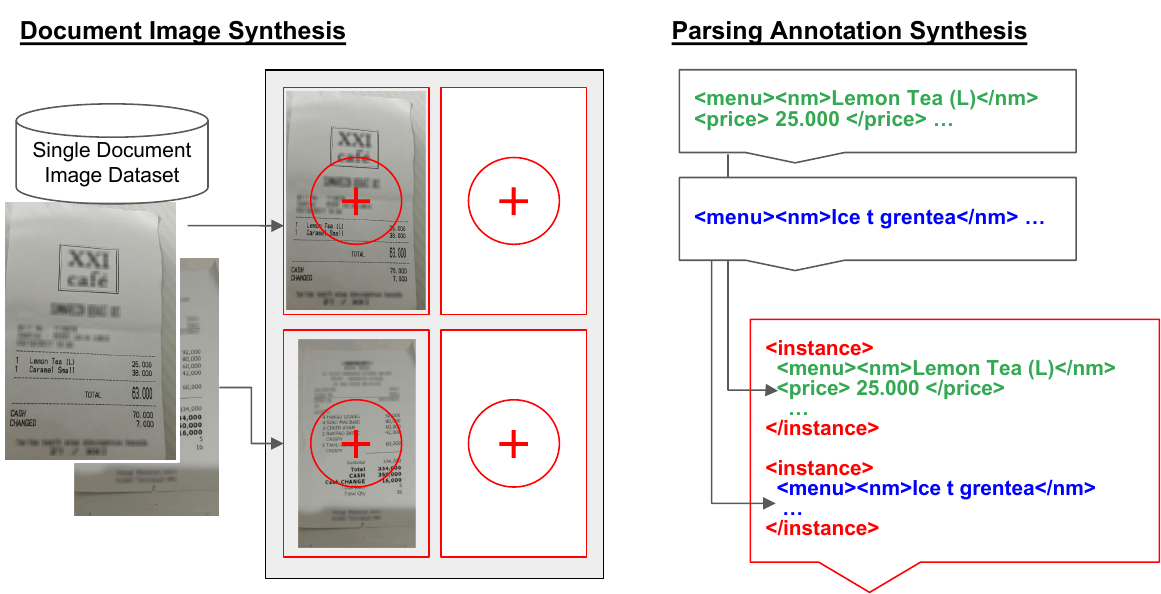}
\caption{
\textbf{Procedure for Dynamically Generating Multi-Document Image Datasets from Single-Document Image Datasets.} 
To generate multi-document images, a random number of document images are arranged and pasted onto a single canvas. 
Simultaneously, the parsing annotations for each document image are concatenated to create an annotation for the multi-document image.
The key idea here is the insertion of special tokens \texttt{<instance>} and \texttt{</instance>} to enclose each parsing annotation when concatenating them, to ensure they are distinguishable even in a text sequence format.
}
\label{fig:synthesis}
\end{figure}

Furthermore, we utilize a document image and annotation synthesis technique to create multi-document image dataset from existing document image datasets. The document image synthesis and the annotation synthesis are conducted based on the copy-and-paste operation and the concatenation operation, respectively. The details of the method are described in Fig.~\ref{fig:synthesis}. This eliminates the need to construct a dedicated multi-document image dataset.

\newpage
\subsection{Experiment}
We trained and evaluated the CREPE model on synthesized dataset from CORD~\cite{park2019cord} dataset.
Each multi-document image was created by randomly selecting 1 to 4 images from the CORD dataset. In the training phase, to increase the variation in data and enhance the learning effect, this process is carried out for each batch. In the testing phase, multi-document images are produced using each of the 100 images from the CORD dataset exactly once, resulting in a total of 32 multi-document images.

The results, shown in Table~\ref{tab:method_comparison}, demonstrate that our CREPE model is capable of parsing multi-document images in bulk with only a minimal decrease in F1-score, which drops merely from 84.7 to 84.2 when compared to the parsing of individual single documents.
In this experiment, it is important to note that even if the resolution of the input images remains the same, the size of the documents within the synthesized multi-document images is reduced to one-fourth of what it is in single-document images. Therefore, it is desirable to increase the resolution of the input images for multi-document cases, which can be considered one of the limitations.

\begin{table}[h]
\centering
\caption{Comparison of Document Parsing Methods}
\label{tab:method_comparison}
\begin{tabular*}{\textwidth}{@{\extracolsep{\fill}}lcccccc}
\toprule
Method & F1 & Docs per Image & Total Images & Total Docs & Resolution & Epoch\\
\midrule
CREPE {\small single-doc}& 84.4 & 1 & 100 & 100 & 1280*960 & 150\\
CREPE {\small multi-doc}& 77.6 & 1--4 & 32 & 100 & 1280*960 & 150\\
CREPE {\small multi-doc}& 83.1 & 1--4 & 32 & 100 & 1920*1600 & 150\\
CREPE {\small multi-doc}& 84.2 & 1--4 & 32 & 100 & 1920*1600 & 300\\
\bottomrule
\end{tabular*}
\end{table}

\begin{figure}[p]
\centering
\includegraphics[width=\textwidth]{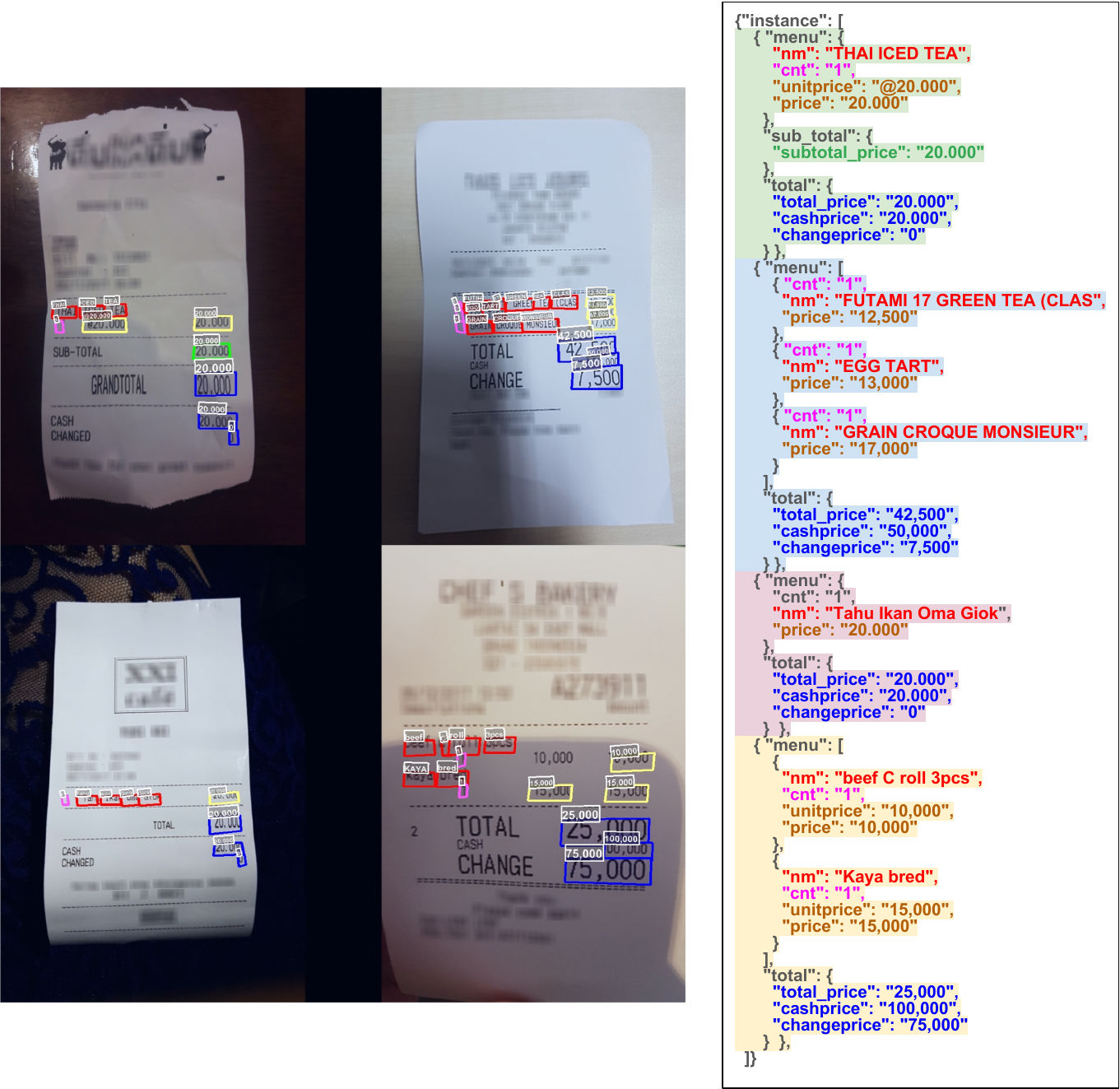}
\caption{
\textbf{Parsing Result for Multi Document Image.} 
Even with images containing multiple documents, our proposed method was able to provide the parsing results for each document in a distinguishable manner. The color of bounding boxes represents the semantics of the text.
}
\label{fig:example}
\end{figure}

\newpage
\section{Scene Text Detection and Object Detection.}

\begin{figure}[h]
\centering
\includegraphics[width=\textwidth]{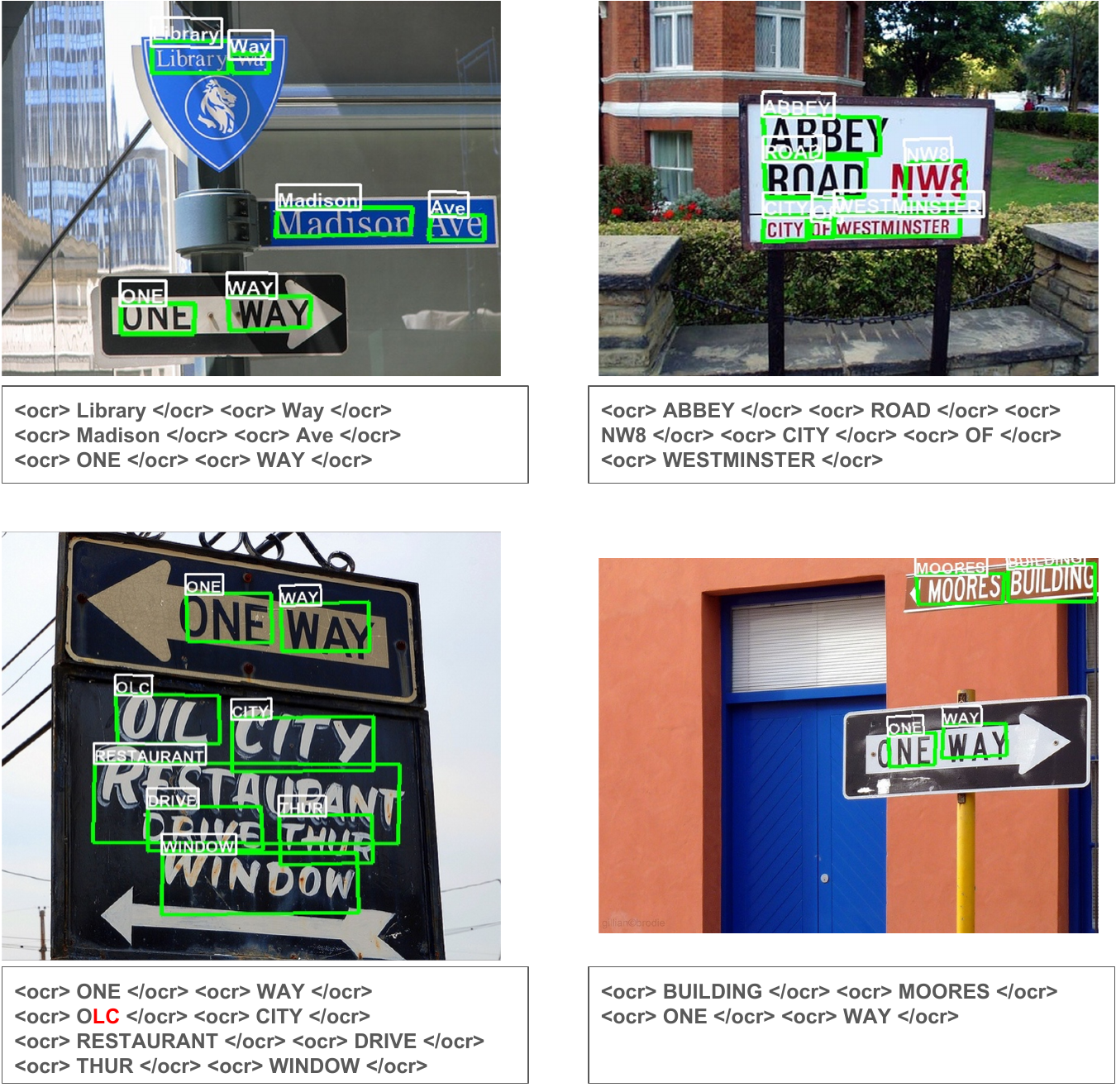}
\caption{
\textbf{Scene Text Detection Result.} 
Our CREPE was generally able to accurately recognize scene text. However, recognition errors occasionally occurred with fonts featuring special designs. The text sequence highlighted in red indicates incorrect strings.
}
\label{fig:scene_text_example}
\end{figure}

\begin{figure}[p]
\centering
\includegraphics[width=\textwidth]{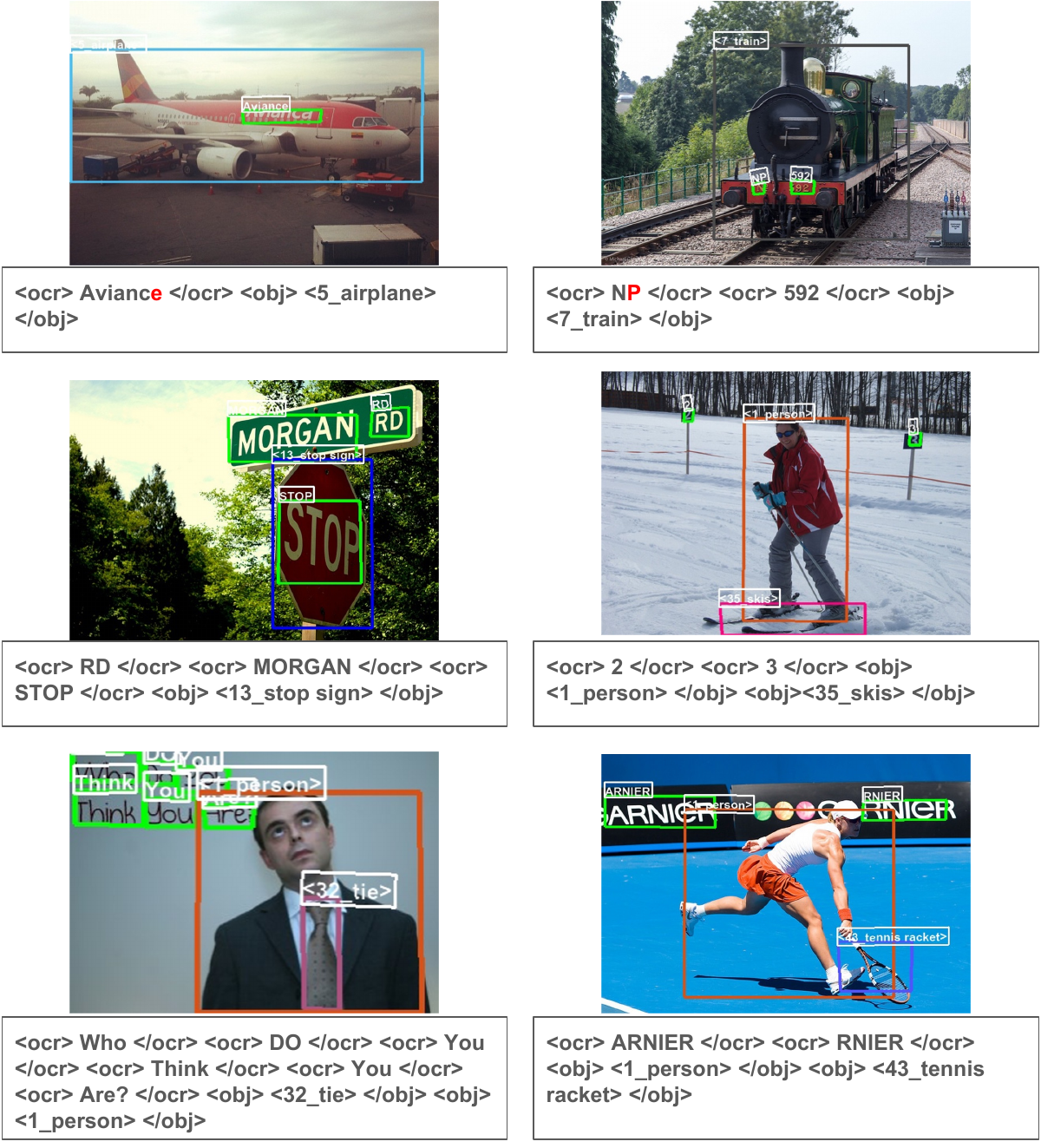}
\caption{
\textbf{Scene Text and Object Detection Result.} 
The text sequence highlighted in red indicates incorrect strings.
}
\label{fig:object_detection_example}
\end{figure}

In this section, we present additional experimental results from Scene Text Detection (STD)~\cite{zhu2016scene} and object detection experiments on COCO 2014 dataset~\cite{lin2014microsoft} conducted in the main text.
As shown in Fig.~\ref{fig:scene_text_example} and Fig.~\ref{fig:object_detection_example}, CREPE demonstrated the ability to recognize the content of signs along with their detection, detect vehicles while simultaneously reading model numbers, and identify text on advertisements behind persons. These results, indicative of its versatility, hold promising implications for a wide range of applications, including video analysis applications and robotics vision.

\vspace{8mm}

\newpage
\section{Acknowledgements}
Some experiments were inspired and conducted based on the product-oriented feedback provided by Ryosuke Kanasugi and Kazuto Ikeda. We express our gratitude for their valuable contributions.

\bibliography{supplement}